\documentclass[a4paper,10pt]{article}

\usepackage{fcds}
\usepackage{graphicx}
\usepackage{float}
\usepackage[labelsep=period]{caption}
\usepackage{enumitem}
\usepackage[utf8]{inputenc}
\usepackage{titlesec}
\titlelabel{\thetitle.\quad}
\usepackage[top=5.5cm, bottom=5cm, left=4cm, right=4cm]{geometry}

\usepackage{booktabs}
\usepackage[noend]{algpseudocode}
\usepackage{tabularx}
\usepackage{algorithm}
\usepackage{subcaption}
\usepackage[colorlinks,
						linkcolor={red},
						citecolor={blue},
						urlcolor={blue},
            pdfauthor={Dariusz Brzezinski},
            pdftitle={Fibonacci and k-Subsecting Recursive Feature Elimination},
            pdfsubject={Feature Selection},
            pdfkeywords={feature selection, wrapper methods, fibonacci method}]{hyperref}

\newcolumntype{C}{>{\centering\arraybackslash}X}
\newcolumntype{R}{>{\raggedleft\arraybackslash}X}
\graphicspath{ {./images/} }

\date{}

\title{\textbf{Fibonacci and $k$-Subsecting Recursive Feature Elimination}}

\author{Dariusz Brzezinski\address{Institute of Computing Science, Poznan University of Technology, Poznan, Poland dariusz.brzezinski@cs.put.poznan.pl}}

\begin{document}
\maketitle

\begin{abstract}
Feature selection is a data mining task with the potential of speeding up classification algorithms, enhancing model comprehensibility, and improving learning accuracy. However, finding a subset of features that is optimal in terms of predictive accuracy is usually computationally intractable. Out of several heuristic approaches to dealing with this problem, the Recursive Feature Elimination (RFE) algorithm has received considerable interest from data mining practitioners. In this paper, we propose two novel algorithms inspired by RFE, called Fibonacci- and $k$-Subsecting Recursive Feature Elimination, which remove features in logarithmic steps, probing the wrapped classifier more densely for the more promising feature subsets. The proposed algorithms are experimentally compared against RFE on 28 highly multidimensional datasets and evaluated in a practical case study involving 3D electron density maps from the Protein Data Bank. The results show that Fibonacci and $k$-Subsecting Recursive Feature Elimination are capable of selecting a smaller subset of features much faster than standard RFE, while achieving comparable predictive performance.
\end{abstract}

\begin{keywords}
feature selection, filter-wrapper methods, recursive feature elimination, Fibonacci search
\end{keywords}

\section{Introduction}
\label{sec:introduction}

Data mining is an iterative process filled with trial and error. The main steps of this process usually include~\cite{journals/ker/MariscalMF10}: business understanding, feature engineering, pattern discovery, and evaluation. Out of all these stages, feature engineering is disputably the most difficult and time-consuming task because it is domain-specific. During feature design, data miners repeatedly gather data, integrate them, clean them and preprocess them, checking with each iteration whether the newly exposed attributes improve the learning model~\cite{DomingosUsefulThings}. As a result of intensive feature engineering, one occasionally faces a scenario where a learning model has difficulties with incorporating all the designed features. In such cases, a feature selection step is needed.

Feature selection is a data mining technique for reducing data dimensionality by removing redundant or irrelevant features from the dataset~\cite{Li2017}. Algorithms that fulfill this task are traditionally divided into wrappers, filters, and embedded methods~\cite{DBLP:journals/jmlr/GuyonE03}. Wrappers utilize the learning algorithm as a black-box to score feature subsets according to their predictive power. Wrappers usually offer feature subsets with better predictive performance than filter methods and, unlike embedded methods, can be easily used with most learning algorithms~\cite{Li2017}. The main drawback of wrapper methods is their computation time, which is strongly connected with the used learning algorithm. As a result, wrappers often rely on heuristics which try to minimize the number of analyzed feature subsets.

Out of several heuristic wrapper selection methods, the Recursive Feature Elimination (RFE)~\cite{RFE} algorithm has received considerable interest. It has been successfully used with various classifiers on biological, chemical, gene-expression, financial, and text data~\cite{RFE,Ha2016,Kalousis2007,Sayes2007}. RFE is a hybrid wrapper-filter algorithm that uses information about feature importance exposed by the wrapped classifier to recursively eliminate the least promising attributes. An important property of this algorithm is that it can be combined with cross-validation to automatically select the optimal number of features to remove. Moreover, to speed up feature selection the user can decide to remove features in larger steps, rather than one attribute at a time. However, larger steps infer larger gaps between consecutively analyzed feature subset sizes. This, in turn, causes RFE to choose too many or too few attributes compared to removing one feature at a time, which negatively affects the number of selected features, and potentially the final learning performance.

In this paper, we put forward two algorithms which attempt to speed up RFE without such loss of selection precision. The first algorithm, called Fibonacci Recursive Feature Elimination (FRFE), searches for the optimal number of features using the Fibonacci method. The second algorithm, called $k$-Subsecting Recursive Feature Elimination ($k$-SRFE), makes $k$ steps over the entire feature range, selects the most promising sub-region and recursively restarts the $k$ steps on the selected sub-region. Both algorithms guarantee the same result as RFE removing one example at a time, when the feature ranking produced by the wrapped classifier remains unchanged and its performance for consecutive feature subsets is a unimodal function. Experiments on 28 benchmark datasets and a case study on crystallographic data show that FRFE and $k$-SRFE are capable of finding smaller subsets of features with comparable or better predictive performance to standard RFE, even when these assumptions might not be fulfilled.

The remainder of this paper is organized as follows: Section~\ref{sec:related} discusses related literature; the proposed Fibonacci and $k$-Subsecting Recursive Feature Elimination algorithms are described in Section~\ref{sec:algorithm}; the experimental comparison against RFE on 28 benchmark datasets is presented in Section~\ref{sec:experiments}; Section~\ref{sec:case} presents a practical application of FRFE and $k$-SRFE to crystallographic data from the Protein Data Bank; and finally conclusions and lines of future research are drawn in Section~\ref{sec:conclusions}.

\section{Related Work}
\label{sec:related}

Feature selection algorithms are usually categorized into wrapper, filter, and embedded methods~\cite{DBLP:journals/jmlr/GuyonE03}. In this study, we focus specifically on a wrapper-filter method called Recursive Feature Elimination (RFE)~\cite{RFE}. RFE is a hybrid of backward feature elimination~\cite{DBLP:journals/ai/KohaviJ97} and feature ranking, which follows a simple iterative procedure:
\begin{enumerate}
	\item Train the classifier;
	\item Compute the feature ranking;
	\item Remove lowest ranked feature.
\end{enumerate}

As the sketched procedure implies, RFE can only be used with algorithms producing feature rankings (e.g. decision trees, SVMs, linear models). Although initially the algorithm was coupled with SVM for selecting features in high-dimensional small sample gene-expression data, it has since been used in a variety of settings with various base models~\cite{Ha2016,Kalousis2007,Li2017,Sayes2007}.

It is worth noting, that the target number of features to select by RFE can be estimated via cross-validation, by picking the number of features with the best mean validation score. However, due to possibly high computational costs, the authors of RFE suggest that in some scenarios it may be more efficient to remove several features at a time, at the expense of possible classification performance degradation~\cite{RFE}. With the purpose of speeding up RFE, Ha and Nguyen recently put forward the Fast Recursive Feature Elimination (Fast RFE) algorithm~\cite{Ha2016}. The authors propose to use Parallel Random Forest as the wrapped classifier, and, thus, save model training time. However, this approach is tied to a concrete learning algorithm and complicates the feature selection process by resorting to distributed cluster computing. Moreover, apart from distributing classifier training, Fast RFE relies on exactly the same iterative process as RFE and removes one feature at a time~\cite{Ha2016}.

In the following section, we present two novel algorithms, which recursively remove features in logarithmic steps. Instead of removing attributes in constant intervals, the algorithms attempt to optimize resource usage by removing larger subsets of the lowest ranked features and analyzing in more detail the most promising feature subsets. In this respect, the proposed methods share the motivation of Bayesian optimization methods and the Auto ML movement~\cite{Bergstra2011,mlrMBO,Hutter2015,Hutter2011}. However, Auto ML algorithms are general purpose methods used mostly for classifier tuning, which do not take into account the fact that performing a removal step in feature elimination affects the results for consecutive steps.

Finally, the methods proposed in this paper are generally inspired by non-linear programming~\cite{NonlinearProgramming}, and the Fibonacci search algorithm in particular~\cite{Fibonacci}. Apart from Fibonacci line search, there are several other optimization methods, such as, bisection,  Newton’s method, or gradient decent~\cite{NonlinearProgramming}, which are related to the tackled problem. However, these methods require derivatives, which cannot be computed for the analyzed, discrete, function of classifier evaluations, and, therefore, they will not be discussed in this paper.

\section{Fibonacci and \texorpdfstring{$k$-Subsecting}{k-Subsecting} Recursive Feature Elimination}
\label{sec:algorithm}

The proposed Fibonacci and $k$-Subsecting Recursive Feature Elimination (FRFE, $k$-SRFE) algorithms are inspired by line search methods~\cite{NonlinearProgramming}. The main goal of FRFE and $k$-SRFE is to find the best number of attributes to select, without having to remove one feature at a time. The basic iterative process of Recursive Feature Elimination (RFE)~\cite{RFE} remains the same, however, the feature subset and the number of features to remove change with each step.

Given a wrapped classifier capable of outputting feature rankings, let us assume that:
\begin{itemize}
	\item[1)]	the feature ranking is stable, i.e., the ranking order does not change upon removing consecutive features;
	\item[2)]	among wrapper mean validation scores there is only one local maximum, i.e. the function of validation scores is unimodal.
\end{itemize}

If these two assumptions were to be met, features can be removed in larger batches without affecting their ranking and the highest mean validation score can be found using a line search method, such as dichotomous search, the golden section method or the Fibonacci method~\cite{NonlinearProgramming}. Given the above, one can select the same number of features as RFE without having to remove one feature at a time. The FRFE and $k$-SRFE algorithms described below rely on these assumptions and offer two alternatives to determining the number of features to select.

FRFE finds the best number of features by performing Fibonacci search within the interval $\left[\mathit{lower}=1, \mathit{upper}=m\right]$, where $m$ is the number of features in the dataset. First, the program calculates the number of iterations required to find the maximum validation score. This is done by solving for the smallest value of $n$ that makes this inequality true: $F_n > (upper - lower)$, where $F_n$ is the $n$-th Fibonacci number. The Fibonacci search involves placing two experiments between $\left[\mathit{lower}, \mathit{upper}\right]$ using subsequent Fibonacci numbers. The function, to be maximized (in this case the function of classifier scores), is evaluated at these two points and the functional values are compared. At each step, we want to keep the feature subset with the higher evaluation score and its corresponding opposite end-point. At the end of the required iterations, one feature subset will be left. The pseudocode for FRFE is presented in Algorithm~\ref{alg:frfe}.

\begin{algorithm}[htbp]
\caption{Fibonacci Recursive Feature Elimination}
\label{alg:frfe}
{\bf Input}: $D$:~$m$-dimensional dataset, $\mathit{cv}$: cross-validation procedure, $Q()$: classifier evaluation measure, $C$: classification algorithm\\ 
{\bf Output}: $S$: set of selected features \\
\begin{algorithmic}[1]
\State $\mathit{lower} \gets 1$, $\mathit{upper} \gets m$, $\mathit{ranks[m]} \gets$ all features                       \Comment Initial search interval
\State $f \gets$ all features, $\mathit{fibs} \gets$ Fibonacci numbers up to the first value greater than $m$
\State $n \gets \mathit{length}(\mathit{fibs}) - 1$
\State $\mathit{ranks}[\mathit{upper}]$, $\mathit{scores}[\mathit{upper}] \gets \mathit{cv}(f, D, Q())$                 \Comment Rank all features in each fold
\State
\State $x1 \gets \mathit{lower} + \mathit{fibs}[n-2]$      \Comment  Evaluate first two feature subsets 
\State $x2 \gets \mathit{lower} + \mathit{fibs}[n-1]$
\State $f_{x1} \gets \mathit{select\_top}(\mathit{ranks}[\mathit{upper}], x1)$
\State $\mathit{ranks}[x1]$, $\mathit{scores}[x1] \gets \mathit{cv}(f_{x1}, D, Q())$
\State $f_{x2} \gets \mathit{select\_top}(\mathit{ranks}[\mathit{upper}], x2)$
\State $\mathit{ranks}[x2]$, $\mathit{scores}[x2] \gets \mathit{cv}(f_{x2}, D, Q())$                
\State $y1 \gets$ mean cv score for $x1$, $y2 \gets$ mean cv score for $x2$
\State       
\While{$\mathit{upper} - \mathit{lower} > 1$} \Comment  Fibonacci search until one feature subset is left
	\State $n \gets n - 1$
	\If{$y1 < y2$}
		\State $\mathit{lower} \gets x1$, $\mathit{upper} \gets \mathit{upper}$
		\State $x1 \gets x2$, $y1 \gets y2$
		\State $x2 \gets \mathit{lower} + \mathit{fibs}[n-1]$
		\If{$\mathit{ranks}[x2] = \emptyset$}
			\State $f_{x2} \gets \mathit{select\_top}(\mathit{ranks}[\mathit{upper}], x2)$
			\State $\mathit{ranks}[x2]$, $\mathit{scores}[x2] \gets \mathit{cv}(f_{x2}, D, Q())$
		\EndIf
		\State $y2 \gets$ mean cv score for $x2$
	\Else
		\State $\mathit{lower} \gets \mathit{lower}$, $\mathit{upper} \gets x2$
		\State $x2 \gets x1$, $y2 \gets y1$
		\State $x1 \gets \mathit{lower} + \mathit{fibs}[n-2]$
		\If{$\mathit{ranks}[x1] = \emptyset$}
			\State $f_{x1} \gets \mathit{select\_top}(\mathit{ranks}[\mathit{upper}], x1)$
			\State $\mathit{ranks}[x1]$, $\mathit{scores}[x1] \gets \mathit{cv}(f_{x1}, D, Q())$
		\EndIf
		\State $y1 \gets$ mean cv score for $x1$
	\EndIf
\EndWhile

\If{$y1 < y2$}
	\State $\mathit{best} \gets x2$
\Else 
	\State $\mathit{best} \gets x1$
\EndIf

\State $S \gets$ remove features without cv by re-analyzing subsets from $m$ to $best$
\end{algorithmic}
\end{algorithm}

The Fibonacci search procedure makes two classifier evaluations at the first iteration and then only one evaluation at each of the subsequent iterations. The number of iterations required to find the optimal number of features is defined by the lowest Fibonacci number greater or equal to m, which grows logarithmically and can be approximated using Binet’s formula~\cite{GoldenRatio}.

It is worth mentioning that among the derivative-free methods that minimize strict quasi-convex functions over a closed bounded interval, the Fibonacci search method is the most efficient in that it requires the smallest number of observations for a given reduction in the length of the searched interval~\cite{NonlinearProgramming}. Moreover, in the context of finding the optimal number of features another advantage of the Fibonacci search over other line search methods is that it works on integers (subsequent Fibonacci numbers) as opposed to real-valued uncertainty intervals. 

SRFE has a similar goal of finding the best number of features by performing a line search. Here, however, instead of performing the minimal number of classifier evaluations we subsect each analyzed interval $k$ times. More precisely, we propose to divide the interval $\left[\mathit{lower}, \mathit{upper}\right]$ using $k$ equally spaced points with $\mathit{step} = \left\lfloor (\mathit{upper} + \mathit{lower})/k\right\rfloor$. Next, the algorithm checks the mean validation score for upper features and the consecutive $k$ points by removing features recursively. Assuming there is only one local maximum, the point with the best mean validation score (best) is closest to the local maximum. If possible, the algorithm updates the search interval accordingly to $\left[\mathit{best} - \mathit{step}, \mathit{best} + \mathit{step}\right]$. The process is continued until the interval has been searched with $\mathit{step} = 1$. The pseudocode for $k$-SRFE is presented in Algorithm~\ref{alg:srfe}.

\begin{algorithm}[!htb]
\caption{$k$-Subsecting Recursive Feature Elimination}
\label{alg:srfe}
{\bf Input}: $D$:~$m$-dimensional dataset, $k$: number of subsecting points, $\mathit{cv}$: cross-validation procedure, $Q()$: classifier evaluation measure, $C$: classification algorithm\\ 
{\bf Output}: $S$: set of selected features \\
\begin{algorithmic}[1]
\State $\mathit{lower} \gets 1$, $\mathit{upper} \gets m$, $\mathit{ranks[m]} \gets$ all features                       \Comment Initial search interval
\State $f \gets$ all features, $\mathit{step} \gets \left\lfloor(\mathit{upper} + \mathit{lower})/k\right\rfloor$
\If{$\mathit{step} = 0$}
	\State $\mathit{step} \gets 1$
\EndIf
\State $\mathit{ranks}[\mathit{upper}]$, $\mathit{scores}[\mathit{upper}] \gets \mathit{cv}(f, D, Q())$ \Comment Rank all features in each fold
\State
\While{$\mathit{step} > 0$}
	\State $\mathit{prev\_mid} \gets \mathit{upper}$ 
	\State $\mathit{mid} \gets \mathit{upper} - \mathit{step}$

  \While{$\mathit{mid} > \mathit{lower} - \mathit{step}$} \Comment Recursively remove features
		\State $\mathit{mid} \gets \max(\mathit{mid}, \mathit{lower})$
		\State $f \gets \mathit{select\_top}(\mathit{ranks}[\mathit{prev\_mid}], \mathit{mid})$
		\State $\mathit{ranks}[\mathit{mid}], \mathit{scores}[\mathit{mid}] \gets \mathit{cv}(f, D, Q())$ \Comment Store cv-scores and rankings
		\State $\mathit{prev\_mid} \gets \mathit{mid}$
		\State $\mathit{mid} \gets \mathit{prev\_mid} - \mathit{step}$
	\EndWhile
	\State $best \gets$ number of features with highest mean cv-score from last $k$ steps
	\State
	\State \textbf{if} $\mathit{best} - \mathit{step} > 1$ \textbf{then} $\mathit{lower} \gets \mathit{best} - \mathit{step}$
	\State \textbf{else} $\mathit{lower} \gets 1$
	\State
	\State \textbf{if} $\mathit{best} + \mathit{step} < m$ \textbf{then} $\mathit{upper} \gets \mathit{best} + \mathit{step}$
	\State \textbf{else} $\mathit{upper} \gets m$
	\State
	\State \textbf{if} $\mathit{step} > 1$ \textbf{and} $\left\lfloor(\mathit{upper} - \mathit{lower})/k\right\rfloor = 0$ \textbf{then} $\mathit{step} \gets 1$
	\State \textbf{else} $\mathit{step} \gets \min(\left\lfloor (\mathit{upper} - \mathit{lower})/k\right\rfloor, \mathit{step} - 1)$
\EndWhile
\State $S \gets$ remove features without cv by re-analyzing subsets from $m$ to $best$
\end{algorithmic}
\end{algorithm}

\begin{figure}[htb]
    \centering
    \begin{subfigure}[b]{0.4\textwidth}
        \includegraphics[width=\textwidth]{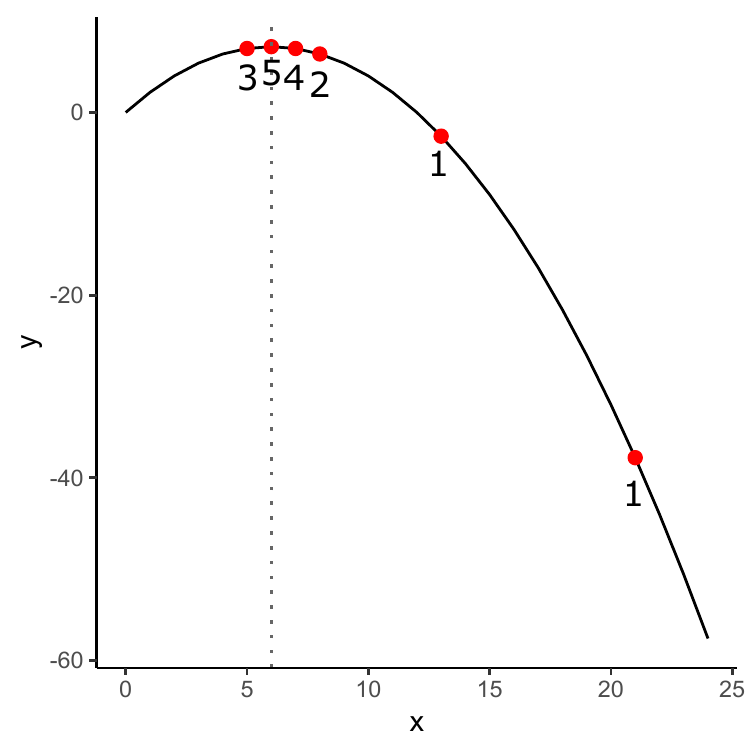}
        \caption{FRFE}
    \end{subfigure}
    ~
    \begin{subfigure}[b]{0.4\textwidth}
        \includegraphics[width=\textwidth]{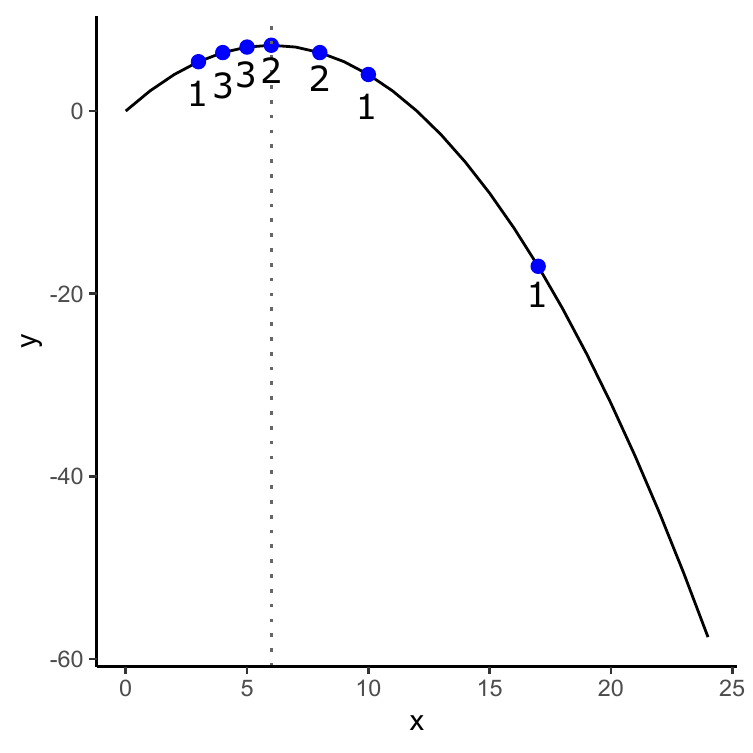}
        \caption{$k$-SRFE}
    \end{subfigure}
		\caption{Comparison of FRFE and 3-SRFE on function $f(x) = -0.2x^2 + 2x$. Function $f(x)$ is used as a hypothetical unimodal function of classifier performance ($y$) for consecutive feature subsets ($x$). Points depict feature subsets probed by FRFE (left) and 3-SRFE (right). Numbers next to points show the algorithm iteration a at which a feature subset was probed.}
		\label{fig:comparison}
\end{figure}

In contrast to FRFE, $k$-SRFE makes more evaluations than is required to find the maximum of a unimodal function. However, if the classifier’s performance were not strictly unimodal, probing more feature subsets gives $k$-SRFE a chance to find a better solution. Figure~\ref{fig:comparison} compares FRFE and $k$-SRFE (with $k = 3$) on an example unimodal function.

Apart from using Fibonacci line search and $k$-subsecting, one could use other optimization methods, such as, bisection,  Newton’s method, or gradient decent~\cite{NonlinearProgramming}. However, these methods require derivatives, which cannot be computed for the analyzed, discrete, function of classifier evaluations.

Even though assumptions 1) and 2) mentioned at the beginning of this section are very strong and may not be met in practical scenarios, during recursive feature elimination the validation score typically rises up to a point when further elimination results in a drastic drop in predictive accuracy. This observation makes FRFE and $k$-SRFE viable heuristics to traditional recursive feature elimination. The following sections verify how the proposed algorithms perform in practical scenarios. Although, in this study we focus on classification problems, the proposed algorithms can be also used for regression. Finally, it is worth noting that, contrary to standard RFE, the user may wish to stop FRFE or $k$-SRFE earlier to obtain an interval of promising feature subsets.

\section{Experimental Evaluation}
\label{sec:experiments}
In this section, we experimentally compare the proposed FRFE and $k$-SRFE algorithms with RFE~\cite{RFE}, on 28 high-dimensional classification datasets. The algorithms are analyzed in terms of the number of selected features, standard deviation of the number of features, predictive performance, and processing time.

\subsection{Setup}
The aim of the experiments is to compare FRFE and $k$-SRFE with RFE parameterized to use the same number of steps. $k$-SRFE will be analyzed for $k \in \left\{3, 5, 10\right\}$, thus, producing the following algorithm pairs for comparison:
\begin{itemize}
	\item FRFE with RFE$_F$,
	\item 3-SRFE with RFE$_3$,
	\item 5-SRFE with RFE$_5$,
	\item 10-SRFE with RFE$_{10}$.
\end{itemize}

For each dataset, RFE$_F$ has the same number of feature elimination steps as FRFE, RFE$_3$ as 3-SRFE, etc. For example, if FRFE analyzed 12 feature subsets before finding the best one on a dataset with $m$=240 features, then RFE$_F$ will also perform 12 evaluations uniformly distributed between $m$ and 1 feature $(240, 220, 200, 180, \ldots)$. In cases where it was impossible to produce the same number of steps between $m$ and 1, RFE uses one additional step. We note that RFE does not offer any mechanism for determining a good feature removal step size. The ability to automatically determine the number of features to be removed at each step is an inherent asset of FRFE and $k$-SRFE. Nevertheless, we compare our proposal against RFE, as if the number of feature evaluations would be known to RFE a priori. All the recursive elimination algorithms used stratified 5-fold cross-validated accuracy when assessing a given number of features.

Four classifiers were used for comparing feature selection methods: Logistic Regression (LR), SVM (SVM), Random Forest (RF), and Gradient Boosting Machines (GBM). To ensure with high probability that Random Forest uses all the features, it was tuned to have 30 trees and use 30\% of attributes for each tree ($3 \times 10^{-15}$ chance of not selecting an attribute to any of the trees). SVM used a linear kernel and was limited to 1000 iterations, whereas all the remaining parameters were left with default scikit-learn~\cite{scikit-learn} values. The features were normalized using min-max [0-1] scaling calibrated on the training folds. Predictive performance was evaluated using: classification accuracy, the kappa statistic, macro-averaged recall, and G-mean~\cite{Japkowiczbook}. Results for each dataset were averaged over 10 cross-validation runs.

All the algorithms were implemented in Python using the scikit-learn library~\cite{scikit-learn}.\footnote{Source codes, datasets, and scripts available at: \url{https://github.com/dabrze/subsecting_rfe}} For Gradient Boosting Machines, we used Microsoft’s LightGBM package.\footnote{\url{https://github.com/Microsoft/LightGBM}} Experiments were conducted on an Amazon EC2\footnote{\url{https://aws.amazon.com/ec2/}} r4.8xlarge virtual machine equipped with 32 vCPUs and 244 GB of RAM.

\subsection{Datasets}
The experimental comparison was performed on 28 benchmark datasets used in a recent feature selection survey by Li et al.~\cite{FeatureSelectionACM}. Table~\ref{tab:datasets} summarizes each dataset.

\begin{table}[!htbp]
\centering
	\caption{Dataset characteristics.}
	\label{tab:datasets}
\begin{tabularx}{\textwidth}{@{}lX@{}X@{}R@{}R@{}R@{}}
\toprule
Dataset     & Type              & Feature types & \# Features & \# Examples & \# Classes \\ \midrule
ALLAML      & Bio               & Continuous    & 7129        & 72          & 2          \\
arcene      & Mass spect. & Continuous    & 10000       & 200         & 2          \\
BASEHOCK    & Text              & Continuous    & 4862        & 1993        & 2          \\
Carcinom    & Bio               & Continuous    & 9182        & 174         & 11         \\
CLL-SUB-111 & Bio               & Continuous    & 11340       & 111         & 3          \\
COIL20      & Image             & Continuous    & 1024        & 1440        & 20         \\
colon       & Bio               & Discrete      & 2000        & 62          & 2          \\
gisette     & Image             & Continuous    & 5000        & 7000        & 2          \\
GLI-85      & Bio               & Continuous    & 22283       & 85          & 2          \\
GLIOMA      & Bio               & Continuous    & 4434        & 50          & 4          \\
Isolet      & Spoken letter     & Continuous    & 617         & 1560        & 26         \\
leukemia    & Bio               & Discrete      & 7070        & 72          & 2          \\
lung        & Bio               & Continuous    & 3312        & 203         & 5          \\
lung\_small & Bio               & Discrete      & 325         & 73          & 7          \\
lymphoma    & Bio               & Discrete      & 4026        & 96          & 9          \\
madelon     & Artificial        & Continuous    & 500         & 2600        & 2          \\
nci9        & Bio               & Discrete      & 9712        & 60          & 9          \\
ORL         & Image             & Continuous    & 1024        & 400         & 40         \\
orlraws10P  & Image             & Continuous    & 10304       & 100         & 10         \\
PCMAC       & Text              & Continuous    & 3289        & 1943        & 2          \\
pixraw10P   & Image             & Continuous    & 10000       & 100         & 10         \\
Prostate-GE & Bio               & Continuous    & 5966        & 102         & 2          \\
RELATHE     & Text              & Continuous    & 4322        & 1427        & 2          \\
TOX-171     & Bio               & Continuous    & 5748        & 171         & 4          \\
USPS        & Image             & Continuous    & 256         & 9298        & 10         \\
warpAR10P   & Image             & Continuous    & 2400        & 130         & 10         \\
warpPIE10P  & Image             & Continuous    & 2420        & 210         & 10         \\
Yale        & Image             & Continuous    & 1024        & 165         & 15         \\ \bottomrule
\end{tabularx}
\end{table}
\clearpage

The datasets represent different data categories, e.g., text, image, biological data. As the survey of Li et al. focuses mainly on filter and embedded feature selectors, the experiments in this study complement results described in~\cite{FeatureSelectionACM}.

\subsection{Results}
Due to the large number of feature selector parameterizations, classifiers, and datasets; detailed tabular results can be found in the online supplementary material.\footnote{\url{http://www.cs.put.poznan.pl/dbrzezinski/publications/FRFE.pdf}} In this and the following section, we present the results by means of selected summaries, plots, and statistical hypothesis tests.

\begin{table}[htbp]
\centering
	\caption{Average Friedman test ranks (lowest best) and $p$-values.}
	\label{tab:friedman}
\begin{tabularx}{\textwidth}{@{}lr@{}R@{}R@{}R@{}R@{}R@{}R@{}R@{\quad}r@{}r@{}}
\toprule
\multicolumn{2}{r}{FRFE}  & RFE$_F$  & 3-SRFE & RFE$_3$  & 5-SRFE & RFE$_5$  & 10-SRFE & RFE$_{10}$ & $p$-val \\ \midrule
\multicolumn{10}{c}{Number of selected features}                                  \\\midrule
GBM & 3.75 & 5.25 & 3.01  & 5.28 & 3.69  & 5.53 & 3.82   & 5.64 & \textbf{0.00}   \\
LR  & 5.14 & 5.91 & 3.60  & 5.32 & 3.96  & 5.35 & 2.39   & 4.30 & \textbf{0.00}   \\
RF  & 5.39 & 4.96 & 3.42  & 5.57 & 3.71  & 4.46 & 3.28   & 5.17 & \textbf{0.00}   \\
SVM & 5.03 & 5.85 & 3.19  & 5.10 & 4.00  & 4.41 & 3.21   & 5.17 & \textbf{0.00}   \\\midrule
\multicolumn{10}{c}{Standard deviation of number of selected features}           \\\midrule
GBM & 3.07 & 3.87 & 3.87  & 4.78 & 4.37  & 5.60 & 4.75   & 5.66 & 0.25   \\
LR  & 3.92 & 5.16 & 3.64  & 5.42 & 3.60  & 5.14 & 3.32   & 5.76 & \textbf{0.00}   \\
RF  & 3.46 & 4.08 & 5.92  & 4.28 & 4.64  & 4.10 & 5.60   & 3.87 & \textbf{0.00}   \\
SVM & 3.17 & 5.39 & 3.75  & 5.60 & 4.00  & 5.57 & 3.42   & 5.07 & \textbf{0.00}   \\\midrule
\multicolumn{10}{c}{Processing time}                                          \\\midrule
GBM & 2.17 & 2.50 & 4.71  & 5.96 & 4.42  & 4.73 & 5.08   & 6.39 & \textbf{0.00}   \\
LR  & 1.91 & 2.41 & 5.23  & 5.73 & 4.01  & 4.91 & 5.08   & 6.69 & \textbf{0.00}   \\
RF  & 1.83 & 1.51 & 7.41  & 4.66 & 4.91  & 3.44 & 6.66   & 5.55 & \textbf{0.00}   \\
SVM & 1.87 & 2.19 & 5.21  & 5.71 & 4.39  & 4.85 & 5.25   & 6.50 & \textbf{0.00}   \\\midrule
\multicolumn{10}{c}{Accuracy}                                               \\\midrule
GBM & 4.64 & 4.35 & 4.78  & 3.76 & 5.03  & 4.23 & 4.78   & 4.39 & 0.34   \\
LR  & 4.30 & 4.82 & 4.26  & 4.83 & 4.32  & 4.55 & 4.78   & 4.10 & 0.85   \\
RF  & 4.66 & 4.75 & 4.03  & 4.41 & 4.05  & 4.96 & 4.91   & 4.21 & 0.67   \\
SVM & 4.64 & 3.82 & 4.44  & 4.33 & 5.58  & 4.00 & 5.14   & 4.01 & \textbf{0.03}   \\\midrule
\multicolumn{10}{c}{Kappa}                                                  \\\midrule
GBM & 4.60 & 4.33 & 4.89  & 3.76 & 5.00  & 4.26 & 4.80   & 4.32 & 0.32   \\
LR  & 4.37 & 4.82 & 4.28  & 4.82 & 4.30  & 4.46 & 4.82   & 4.10 & 0.86   \\
RF  & 4.64 & 4.73 & 4.12  & 4.33 & 4.07  & 5.00 & 4.98   & 4.10 & 0.62   \\
SVM & 4.66 & 3.85 & 4.51  & 4.42 & 5.51  & 4.00 & 5.07   & 3.94 & \textbf{0.05}   \\\midrule
\multicolumn{10}{c}{Macro-averaged recall}                                    \\\midrule
GBM & 4.53 & 4.42 & 4.89  & 3.83 & 5.03  & 4.26 & 4.57   & 4.42 & 0.46   \\
LR  & 4.28 & 4.89 & 4.17  & 4.85 & 4.39  & 4.33 & 4.87   & 4.17 & 0.77   \\
RF  & 4.67 & 4.87 & 4.10  & 4.41 & 3.85  & 4.98 & 4.96   & 4.12 & 0.46   \\
SVM & 4.71 & 3.82 & 4.42  & 4.26 & 5.55  & 4.12 & 5.10   & 3.98 & \textbf{0.04}   \\\midrule
\multicolumn{10}{c}{G-mean}                                                 \\\midrule
GBM & 4.48 & 4.51 & 5.03  & 3.76 & 4.85  & 4.41 & 4.60   & 4.32 & 0.43   \\
LR  & 4.26 & 4.89 & 4.33  & 4.83 & 4.28  & 4.26 & 4.98   & 4.12 & 0.71   \\
RF  & 4.35 & 4.60 & 4.21  & 4.58 & 3.71  & 5.35 & 5.07   & 4.08 & 0.20   \\
SVM & 4.57 & 3.78 & 4.26  & 4.33 & 5.44  & 4.25 & 5.21   & 4.12 & 0.08   \\ \bottomrule
\end{tabularx}
\end{table}

\begin{figure}[htb]
    \centering
    \begin{subfigure}[b]{0.48\textwidth}
        \includegraphics[width=\textwidth]{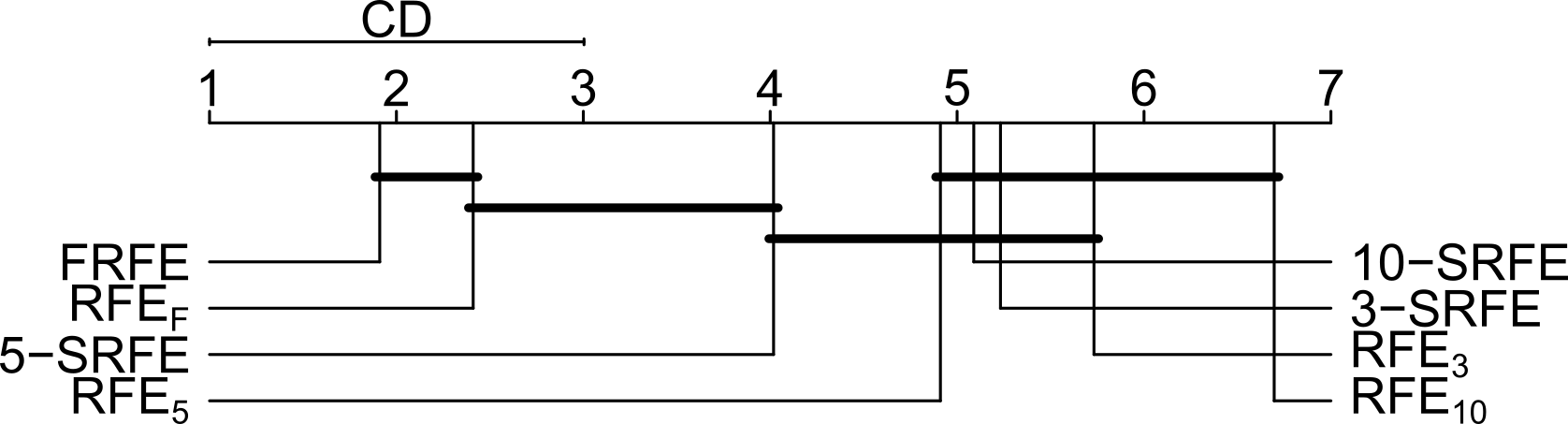}
        \caption{LR}
    \end{subfigure}
    ~
    \begin{subfigure}[b]{0.48\textwidth}
        \includegraphics[width=\textwidth]{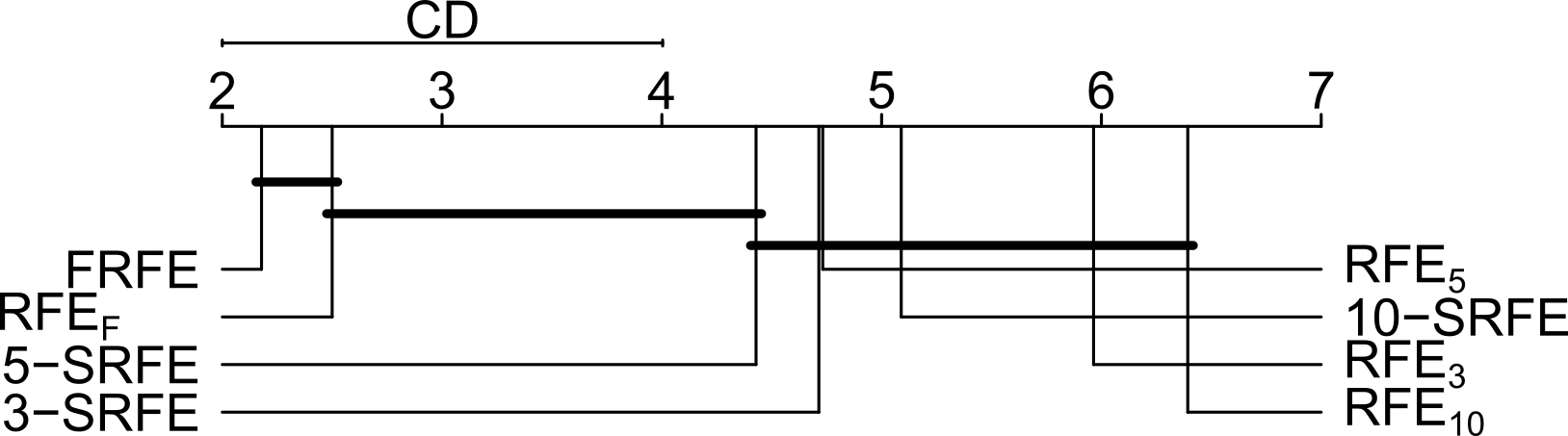}
        \caption{GBM}
    \end{subfigure}

    \begin{subfigure}[b]{0.48\textwidth}
        \includegraphics[width=\textwidth]{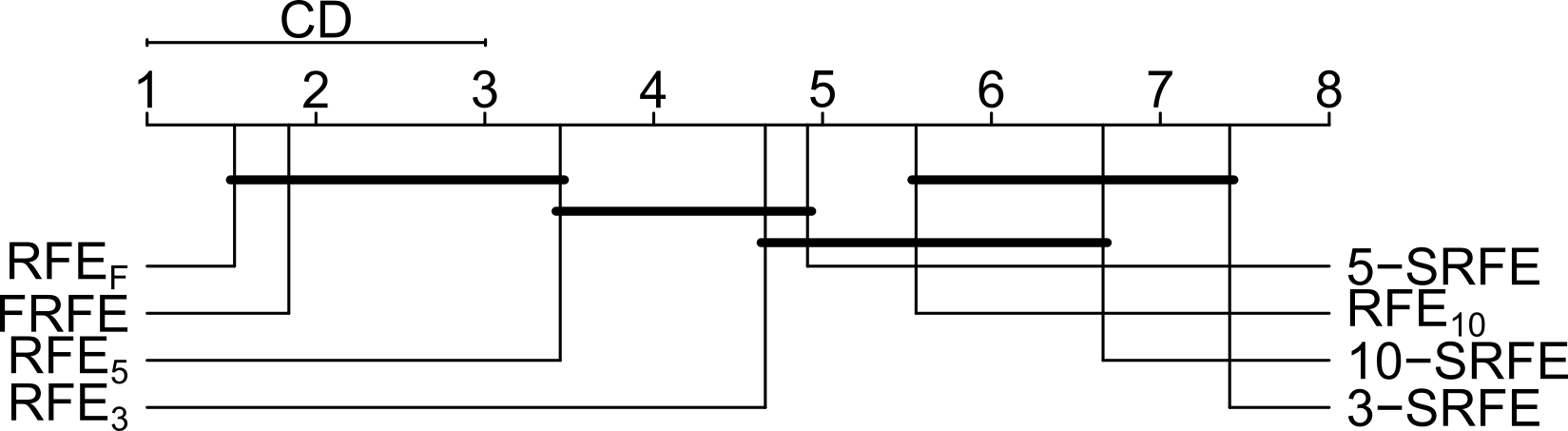}
        \caption{RF}
    \end{subfigure}
		    ~
    \begin{subfigure}[b]{0.48\textwidth}
        \includegraphics[width=\textwidth]{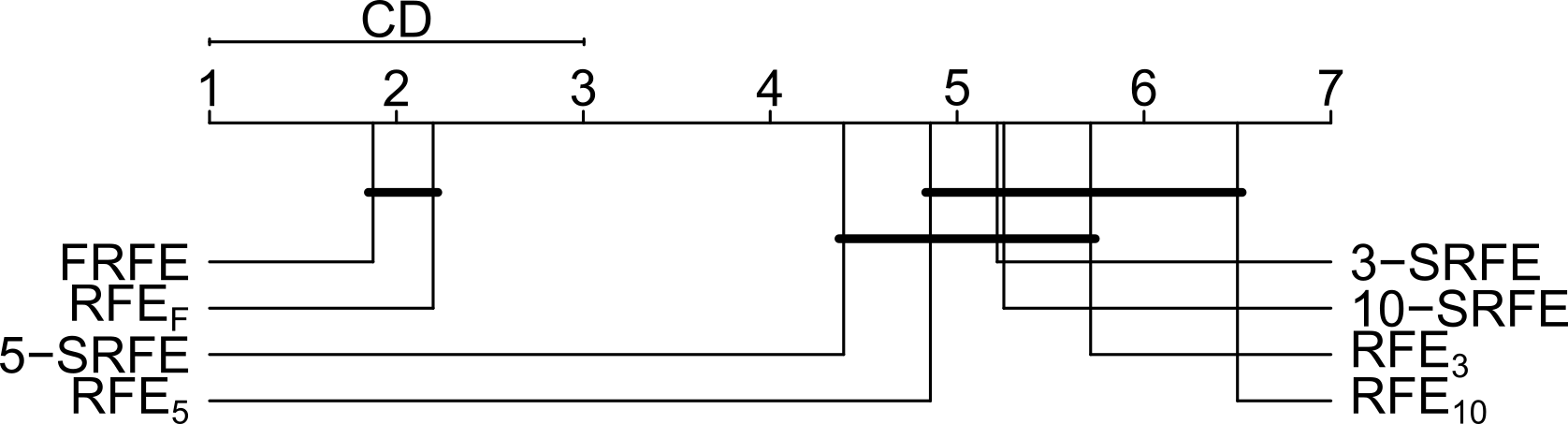}
        \caption{SVM}
    \end{subfigure}
		\caption{Critical distance plots for processing time. Selectors that are not significantly different according to the Nemenyi test (at $\alpha = 0.05$) are connected.}
		\label{fig:friedman-time}
\end{figure}

\begin{figure}[htb]
    \centering
    \begin{subfigure}[b]{0.48\textwidth}
        \includegraphics[width=\textwidth]{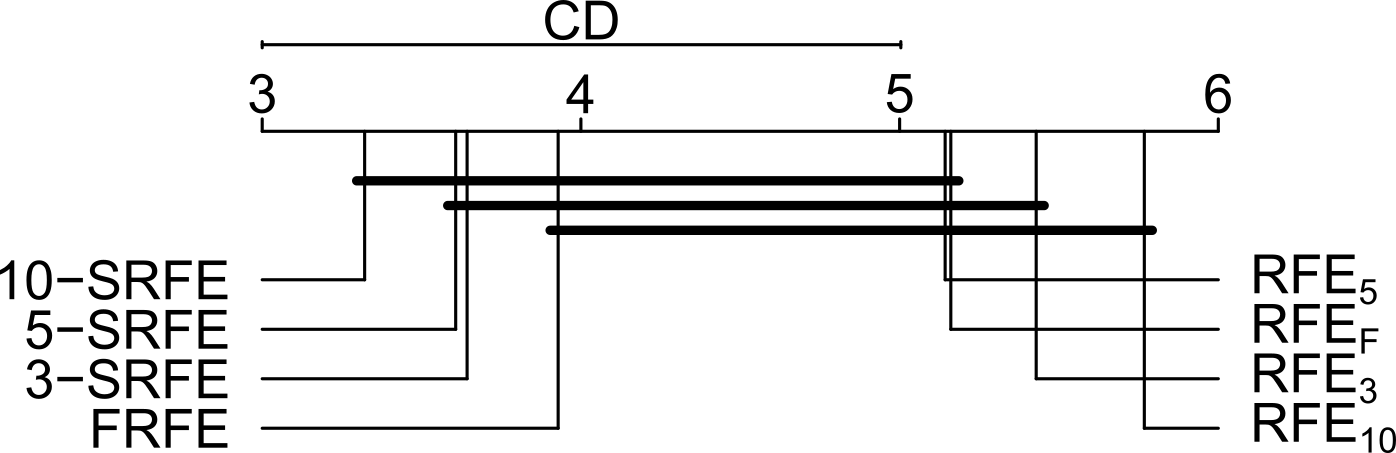}
        \caption{LR}
    \end{subfigure}
    ~
    \begin{subfigure}[b]{0.48\textwidth}
        \includegraphics[width=\textwidth]{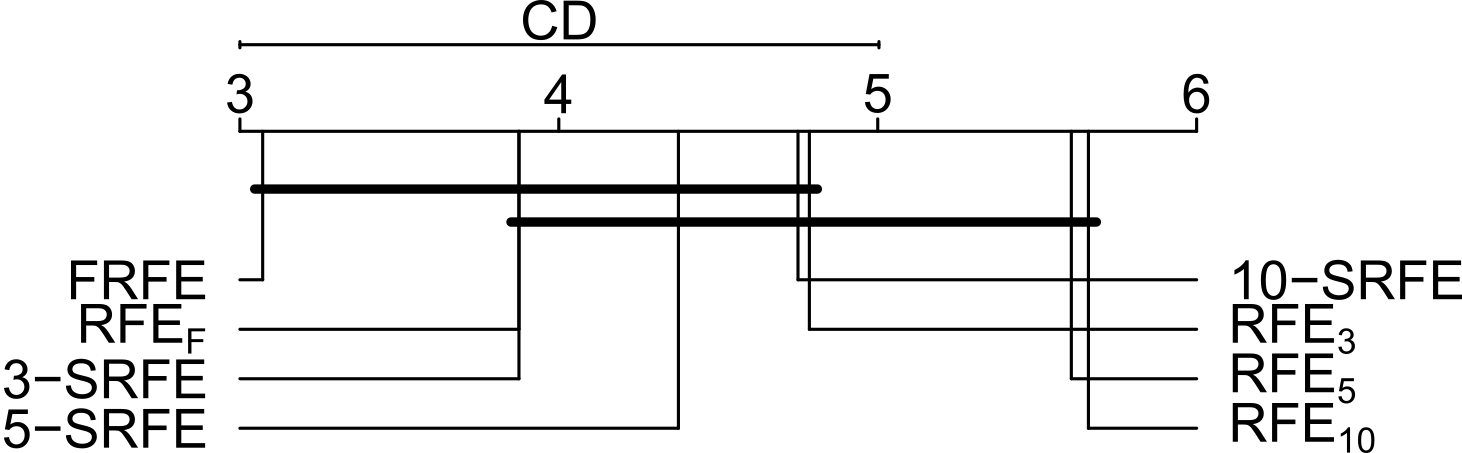}
        \caption{GBM}
    \end{subfigure}

    \begin{subfigure}[b]{0.48\textwidth}
        \includegraphics[width=\textwidth]{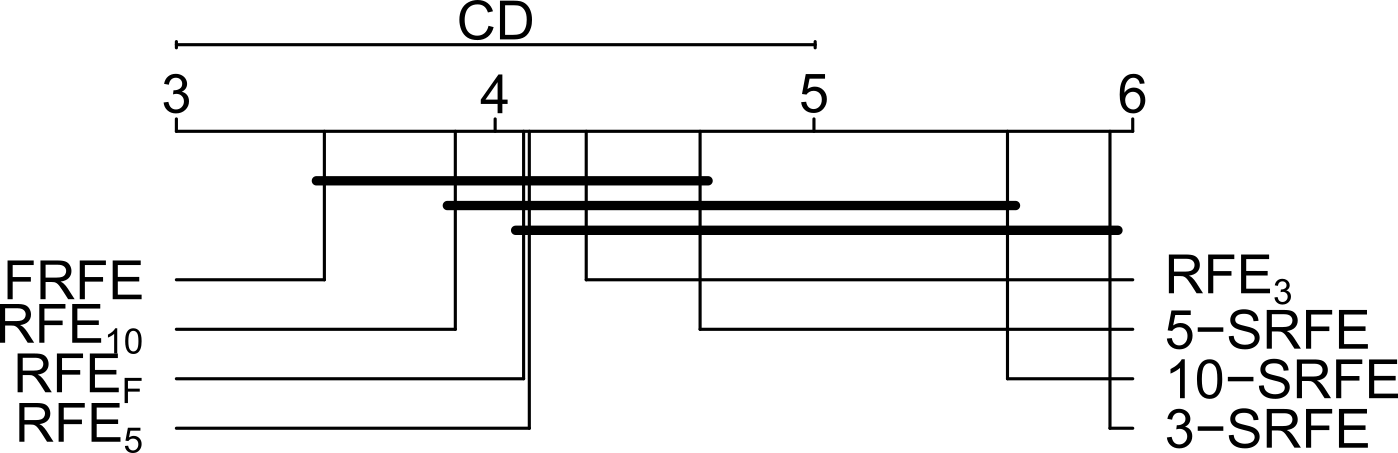}
        \caption{RF}
    \end{subfigure}
		    ~
    \begin{subfigure}[b]{0.48\textwidth}
        \includegraphics[width=\textwidth]{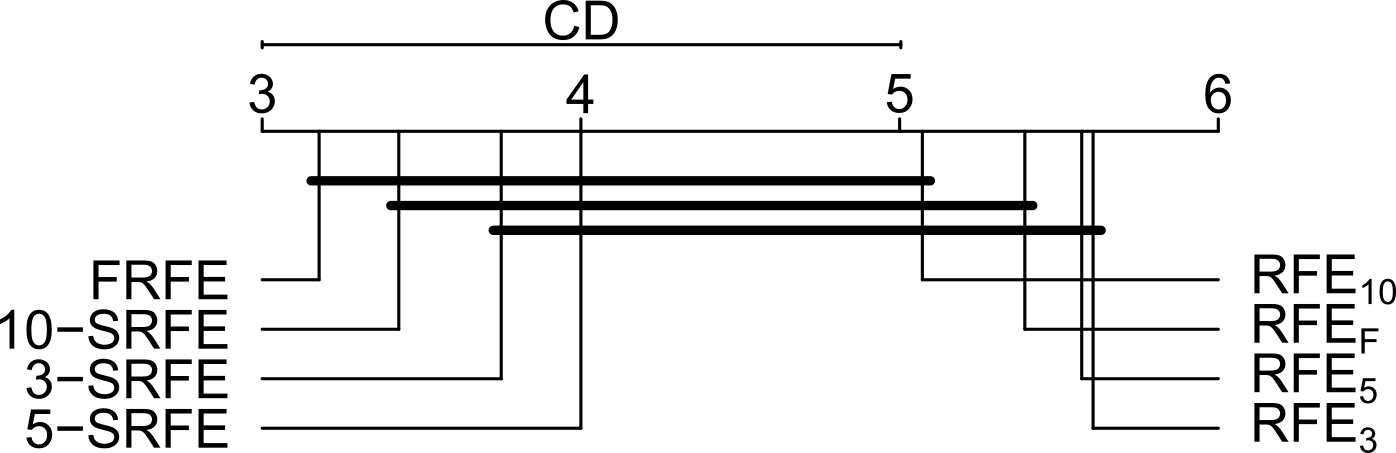}
        \caption{SVM}
    \end{subfigure}
		\caption{Critical distance plots for selected number of features. Selectors that are not significantly different according to the Nemenyi test (at $\alpha = 0.05$) are connected.}
		\label{fig:friedman-features}
\end{figure}

Table~\ref{tab:friedman} presents average Friedman test ranks~\cite{DBLP:journals/jmlr/Demsar06} for each classifier on each tested metric (the lower the rank the better). Setting a significance level $\alpha=0.05$, we can reject the null hypothesis of the Friedman test stating that the performances of the classifiers are equivalent only for the number of selected features, their standard deviation, and processing time. This means that the final predictive performance of classifiers was comparable when using any of the tested feature selectors, yet some them were significantly faster and chose less features than others. The critical distance plots for the Nemenyi post-hoc test~\cite{DBLP:journals/jmlr/Demsar06} (Figures~\ref{fig:friedman-time}~and~\ref{fig:friedman-features}) indicate that FRFE is significantly faster than all the other selectors except RFE$_F$ (which was forced to use the same number of evaluations). Moreover, $k$-SRFE selectors tend to choose smaller feature subsets than their RFE counterparts. 

To verify the significance of differences between pairs of algorithms (RFE$_3$ vs. 3-SRFE, RFE$_5$ vs. 5-SRFE, etc.) we performed the non-parametric Wilcoxon signed ranks test~\cite{DBLP:journals/jmlr/Demsar06} for each pair. Table~\ref{tab:wilcoxon} reports the results of these statistical tests.

\begin{table}[htb]
\centering
	\caption{Wilcoxon signed rank test comparing the proposed algorithms (FRFE, $k$-SRFE) and RFE parametrized to have the same number of steps as the proposed algorithms. Symbol '$>$' denotes superior performance of the proposed algorithm, whereas '$=$' denotes no significant difference at $\alpha = 0.05$.}
	\label{tab:wilcoxon}
\begin{tabularx}{0.9\textwidth}{@{}lRRRR@{}}
\toprule
Hypothesis      & GBM                & LR                 & RF                 & SVM                \\\midrule
\multicolumn{5}{c}{Number of selected features}                                                     \\\midrule
FRFE vs RFEF    & $>  0.003$ & $>  0.048$ & $=  0.866$ & $>  0.023$     \\
3-RFE vs RFE3   & $>  0.000$ & $>  0.011$ & $>  0.007$ & $>  0.010$     \\
5-RFE vs RFE5   & $>  0.002$ & $>  0.023$ & $>  0.045$ & $>  0.035$     \\
10-RFE vs RFE10 & $>  0.003$ & $>  0.002$ & $>  0.008$ & $>  0.000$     \\\midrule
\multicolumn{5}{c}{Standard deviation of number of selected features}   \\\midrule
FRFE vs RFEF    & $=  0.212$ & $>  0.022$ & $=  0.140$ & $>  0.000$     \\
3-RFE vs RFE3   & $>  0.041$ & $>  0.005$ & $=  0.929$ & $>  0.002$     \\
5-RFE vs RFE5   & $>  0.041$ & $>  0.009$ & $=  0.786$ & $>  0.004$     \\
10-RFE vs RFE10 & $>  0.045$ & $>  0.000$ & $=  0.997$ & $>  0.000$     \\\midrule
\multicolumn{5}{c}{Processing time}                                     \\\midrule
FRFE vs RFEF    & $=  0.500$ & $=  0.524$ & $=  0.978$ & $=  0.627$     \\
3-RFE vs RFE3   & $>  0.019$ & $=  0.346$ & $=  1.000$ & $=  0.530$     \\
5-RFE vs RFE5   & $>  0.040$ & $=  0.232$ & $=  1.000$ & $=  0.352$     \\
10-RFE vs RFE10 & $>  0.008$ & $=  0.113$ & $=  0.981$ & $=  0.223$     \\\midrule
\multicolumn{5}{c}{Accuracy}                                            \\\midrule
FRFE vs RFEF    & $=  0.778$ & $=  0.434$ & $=  0.453$ & $=  0.928$     \\
3-RFE vs RFE3   & $=  0.997$ & $=  0.623$ & $=  0.415$ & $=  0.847$     \\
5-RFE vs RFE5   & $=  0.886$ & $=  0.530$ & $=  0.092$ & $=  0.995$     \\
10-RFE vs RFE10 & $=  0.481$ & $=  0.841$ & $=  0.928$ & $=  0.839$     \\
\bottomrule
\end{tabularx}
\end{table}

The pairwise comparisons of different parameterizations of RFE and the proposed methods confirm the observations from the averaged Friedman ranks. FRFE and $k$-SRFE select significantly fewer features with smaller standard deviation than their RFE counterparts. It has also been noticed that Random Forest produces the least stable feature rankings. There were no significant differences in terms of accuracy. Due to the way the experiments were prepared, processing time in each pair of algorithms is very similar.

Supplementary Tables S1-S3 present the number of features, processing time, and accuracy for SVM; detailed results for the remaining classifiers are also available in the supplementary material.\footnote{\url{http://www.cs.put.poznan.pl/dbrzezinski/publications/FRFE.pdf}} Looking at Supplementary Table S1, one can notice that the number of features selected by SVM with $k$-SRFE is not only lower in terms of statistical significance, but also usually much lower in terms of raw values (e.g. 137 features selected by 10-SRFE compared to 1316 by RFE$_{10}$). Similarly, Supplementary Table S2 shows that the processing time of FRFE and its equivalent RFEF are usually much faster than $k$-SRFE and RFE$_k$ selectors which require more classifier evaluations. Finally, in terms of classification accuracy, presented in Supplementary Table S3, there is no clear winner.

\begin{figure}[htb]
	\centering
		\includegraphics[width=\textwidth]{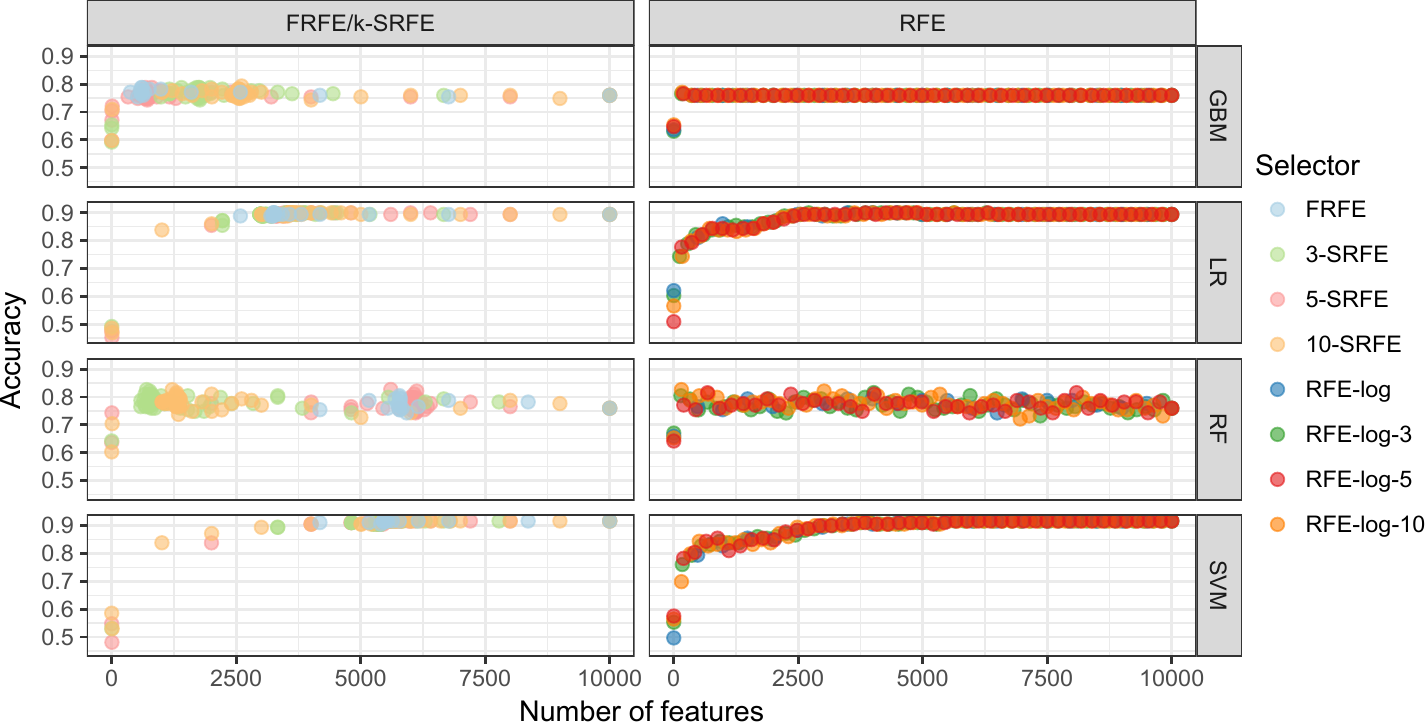}
	\caption{Mean validation scores for a single experiment fold during feature elimination on the arcene dataset; plot best viewed online in color.}
	\label{fig:arcene}
\end{figure}

Figure~\ref{fig:arcene} presents mean validation scores for a single experiment fold on the arcene dataset. The plot depicts characteristic behavior of the tested feature selection methods, as well as the classifiers they were tested on. By looking at the left and right columns, one can compare distributions of the tested feature subsets. FRFE and $k$-SRFE (left column) focus on the most promising features found in previous iterations, whereas RFE (right column) eliminates features at regular intervals. Moreover, it can be noticed that FRFE and $k$-SRFE concentrate their classifier evaluations around the number of features with the highest validation score for each classifier. The plot also confirms that Random Forest, due to its randomness, produces the least stable feature rankings. Conversely, SVM, LR, and GBM produce fairly stable feature rankings.

It is worth noticing that the benchmark datasets proposed by~\cite{FeatureSelectionACM} were chosen to test feature selection algorithms mainly with the goal of improving model size and comprehensibility, not its predictive performance. On most datasets, using all the supplied features ensures very good predictions. On the other hand,  RFE, FRFE, $k$-SRFE, as well as other wrapper methods are mostly used with the purpose of improving model predictions. Therefore, in the following section we apply and compare the algorithms on a real-world dataset with noisy and highly correlated features, which impede classification performance.

\section{Case Study on 3D Electron Density Maps}
\label{sec:case}

Our case study concerns selecting features for detecting ligand structures from macro-molecular electron density maps generated by X-ray crystallography. Given a 3D map of experimental electron density, chemists and biologists model the structure of crystallized molecules, usually to extract important information about a protein’s function. With existing model building software~\cite{Buccaneer,Perrakis1999,RESOLVE}, the regions of macromolecular structure corresponding to polypeptide or polynucleotide chains can be built with high accuracy and speed. On the other hand, small-molecule ligands are usually modeled manually, and their correct identification often requires good judgment and expertise. The presented case study is part of an ongoing project~\cite{CMB} targeted at designing an approach that uses machine learning algorithms to identify ligands in electron density maps. A crucial part of the system lies in feature engineering and selection.

The data presented in this paper were created from 157,953 electron density map fragments of the 20 most popular ligands with at least 2 non-H atoms stored in the Protein Data Bank (PDB)~\cite{PDB}: SO4, GOL, EDO, NAG, PO4, ACT, DMS, HEM, FAD, PEG, MLY, NAD, FMT, NAG-NAG, MPD, NAP, MES, FMN, ADP, NO3. Each example is described by 424 attributes, such as volume, electron density sum, 3D moment invariants, and Zernike invariants~\cite{Gunasekaran2009,Hattne2011,Sommer2007}.

The studied data contain characteristic difficulty factors. First, the classes are severely imbalanced: SO4 has 48,490 examples, whereas NO3 has only 1301 examples. Second, some of the features describing the ligands are strongly correlated with each other, e.g., invariants of Zernike polynomials of similar degrees have correlated values. Moreover, crystallographic images need a cutoff density threshold to determine ligand positions and shapes. By raising this threshold, peaks in an electron density map start to disappear. Therefore, feature values were calculated for three different cutoff values (sigma thresholds), causing high correlation between the same features calculated for different thresholds. Finally, electron density maps can be obtained in different resolutions, depending on the quality of the analyzed crystal. This introduces varying amounts of noise to the images and occasionally causes experts to incorrectly assign ligand labels~\cite{Pozharski2013}.

Table~\ref{tab:case} presents the average evaluation metrics for a wrapped Gradient Boosting Machine, which offered the highest accuracy on such noisy data. As with the benchmark datasets, results were obtained using stratified 10-fold cross-validation.

\begin{table}[htb]
\centering
	\caption{Evaluation metrics on the PDB dataset, averaged over 10 cross validation folds; best value in bold.}
	\label{tab:case}
\begin{tabular}{@{}lrrrrr@{}}
\toprule
                       & FRFE   & 3-SRFE & 5-SRFE & 10-SRFE & All  \\ \midrule
Num. of selected features & 344    & \textbf{271}    & 291     & 291   & (424)     \\
SD of number of selected features & \textbf{62}     & 95     & 125     & 100   & -     \\
Processing time {[}s{]}           & \textbf{30575}  & 54034  & 48650   & 56326 & -   \\
Accuracy               & 0.675  & 0.676  & \textbf{0.676}   & 0.676 & 0.676 \\
Kappa                  & 0.603  & 0.603  & \textbf{0.604}  & 0.603 & 0.603 \\
Macro recall           & 0.564  & 0.565  & \textbf{0.566}   & 0.565 & 0.565 \\
G-mean                 & 0.457  & 0.457  & \textbf{0.458}   & 0.457 & 0.457 \\ \bottomrule
\end{tabular}
\end{table}

For the case study, all the proposed feature selectors achieved similar predictive performance. The only distinguishing factor is the processing time, which was markedly lower for FRFE compared to $k$-SRFE. Low processing time of FRFE was achieved even though it selected, on average, the largest number of features and, therefore, concentrated on evaluating the most populous feature subsets. It is also worth mentioning that for this dataset the estimated running time of standard RFE, calculated by linearly interpolating the number of steps of 10-SRFE, was 555,322 s (154 h). This is almost twenty times more than the time required to run FRFE.

By analyzing common features selected between each evaluation fold, we have highlighted the importance and stability of several features. The results show that attributes based on PCA eigenvalues of electron density fragments are important as they give a rotation-invariant description of the ligand. Moreover, features that report the ligand’s diameter and the differences (deltas) between the estimated number of electrons for consecutive contour levels are also common between FRFE and $k$-SRFE runs. Finally, the number of modeled biopolymer atoms (O, N, C) in the proximity of an electron density fragment add important chemical information to the ligand’s description.

The presented data is part of a larger project aiming at automatically detecting ligands from un-modeled electron density fragments of crystallographic images~\cite{CMB}. The project’s image processing pipeline is constantly being refined and the features describing the ligands change regularly. As a result, features must be repeatedly re-evaluated and re-selected. As the final goal is to distinguish between 200 types of ligands described in over 290,000 examples, feature selection is crucial in this application to speed up model training. In practice, standard RFE is computationally too expensive for such data, and the proposed methods offer a much faster alternative.

\section{Conclusions}
\label{sec:conclusions}

In this paper, we presented two new feature selection algorithms: Fibonacci (FRFE) and $k$-Subsecting Feature Elimination ($k$-SRFE). The proposed algorithms rely on line search methods to determine the best number of features to select. With each iteration the algorithms analyze the feature subsets in more detail, concentrating on smaller feature intervals. As a result, the proposed algorithms automatically adjust the feature search step, offer a significant speed up compared to RFE, and can be stopped before completion to obtain an interval of promising feature subsets. Experiments on 28 benchmark datasets and a case study involving 3D crystallographic data have shown that FRFE chooses a small feature subset very quickly, which could be of great value in applications such as gene selection. On the other hand, $k$-SRFE was capable of achieving comparable or better accuracy using less features than RFE parametrized to use the same number of classifier evaluations.

As future research, we plan to experiment with alternative ways of analyzing the function of feature subset performance. This can be done, for example, by employing surrogate functions that take into account the interaction between consecutive steps in feature elimination. Moreover, we plan to study the performance of the proposed algorithms for regression problems and the impact of using evaluation metrics other than classification accuracy during validation. Finally, the automatic identification of ligands in electron density maps is an ongoing project that constitutes a complex testbed for feature selection and feature extraction methods.

\section*{Data Availability}
To promote reproducible research, the methods described in this paper are available at \url{https://github.com/dabrze/subsecting_rfe}. The repository includes experimental scripts, datasets, and result files. The data display items presented in this manuscript can be reproduced using scripts provided in the aforementioned source code repository.

\section*{Acknowledgment}
The author would like to thank Marcin Kowiel for insightful comments on a draft of this paper. 

\bibliographystyle{fcds_abbrv}
\bibliography{fibonacci_rfe}

\begin{thebibliography}{10}

\bibitem{NonlinearProgramming}
Bazaraa~M.~S., Sherali~H.~D., and Shetty~C.~M.
\newblock {\em Nonlinear Programming: Theory and Algorithms}.
\newblock Wiley Publishing, 2013.

\bibitem{Bergstra2011}
Bergstra~J.~S., Bardenet~R., Bengio~Y., and K\'{e}gl~B.
\newblock Algorithms for hyper-parameter optimization.
\newblock In Shawe-Taylor~J., Zemel~R.~S., Bartlett~P.~L., Pereira~F., and
  Weinberger~K.~Q., editors, {\em Advances in Neural Information Processing
  Systems 24}, pages 2546--2554. Curran Associates, Inc., 2011.

\bibitem{PDB}
Berman~H.~M., Westbrook~J., Feng~Z., Gilliland~G., Bhat~T.~N., Weissig~H.,
  Shindyalov~I.~N., and Bourne~P.~E.
\newblock The protein data bank.
\newblock {\em Nucleic Acids Research}, 28(1):235--242, 2000.

\bibitem{mlrMBO}
Bischl~B., Richter~J., Bossek~J., Horn~D., Thomas~J., and Lang~M.
\newblock {{mlrMBO}}: {{A Modular Framework}} for {{Model}}-{{Based
  Optimization}} of {{Expensive Black}}-{{Box Functions}}.
\newblock {\em CoRR abs/1703.03373}, 2017.

\bibitem{Buccaneer}
Cowtan~K.
\newblock {The {\it Buccaneer} software for automated model building. 1.
  Tracing protein chains}.
\newblock {\em Acta Crystallographica Section D}, 62(9):1002--1011, Sep 2006.

\bibitem{DBLP:journals/jmlr/Demsar06}
Demsar~J.
\newblock Statistical comparisons of classifiers over multiple data sets.
\newblock {\em Journal of Machine Learning Research}, 7:1--30, 2006.

\bibitem{DomingosUsefulThings}
Domingos~P.~M.
\newblock A few useful things to know about machine learning.
\newblock {\em Communications of the {ACM}}, 55(10):78--87, 2012.

\bibitem{Hutter2015}
Feurer~M., Klein~A., Eggensperger~K., Springenberg~J., Blum~M., and Hutter~F.
\newblock Efficient and robust automated machine learning.
\newblock In Cortes~C., Lawrence~N.~D., Lee~D.~D., Sugiyama~M., and Garnett~R.,
  editors, {\em Advances in Neural Information Processing Systems 28}, pages
  2962--2970. Curran Associates, Inc., 2015.

\bibitem{Gunasekaran2009}
Gunasekaran~P., Grandison~S., Cowtan~K., Mak~L., Lawson~D.~M., and Morris~R.~J.
\newblock Ligand electron density shape recognition using 3d zernike
  descriptors.
\newblock In {\em Proceedings of the 4th IAPR International Conference on
  Pattern Recognition in Bioinformatics}, PRIB '09, pages 125--136, Berlin,
  Heidelberg, 2009. Springer-Verlag.

\bibitem{DBLP:journals/jmlr/GuyonE03}
Guyon~I. and Elisseeff~A.
\newblock An introduction to variable and feature selection.
\newblock {\em Journal of Machine Learning Research}, 3:1157--1182, 2003.

\bibitem{RFE}
Guyon~I., Weston~J., Barnhill~S., and Vapnik~V.
\newblock Gene selection for cancer classification using support vector
  machines.
\newblock {\em Machine Learning}, 46(1):389--422, 2002.

\bibitem{Ha2016}
Ha~V.-S. and Nguyen~H.-N.
\newblock {FRFE}: Fast recursive feature elimination for credit scoring.
\newblock In {\em Nature of Computation and Communication: Second International
  Conference, ICTCC 2016, Rach Gia, Vietnam, March 17-18, 2016, Revised
  Selected Papers}, pages 133--142, 2016.

\bibitem{Hattne2011}
Hattne~J. and Lamzin~V.~S.
\newblock A moment invariant for evaluating the chirality of three-dimensional
  objects.
\newblock {\em Journal of The Royal Society Interface}, 8(54):144--151, 2011.

\bibitem{Hutter2011}
Hutter~F., Hoos~H.~H., and Leyton-Brown~K.
\newblock Sequential model-based optimization for general algorithm
  configuration.
\newblock In {\em Proceedings of the 5th International Conference on Learning
  and Intelligent Optimization}, LION'05, pages 507--523, Berlin, Heidelberg,
  2011. Springer-Verlag.

\bibitem{Japkowiczbook}
Japkowicz~N. and Shah~M.
\newblock {\em Evaluating Learning Algorithms: A Classification Perspective}.
\newblock Cambridge University Press, 2011.

\bibitem{Kalousis2007}
Kalousis~A., Prados~J., and Hilario~M.
\newblock Stability of feature selection algorithms: A study on
  high-dimensional spaces.
\newblock {\em Knowledge Information Systems}, 12(1):95--116, 2007.

\bibitem{Fibonacci}
Kiefer~J.
\newblock Sequential minimax search for a maximum.
\newblock In {\em Proceedings of the American Mathematical Society, 4}, pages
  502--506. AMS, 1953.

\bibitem{DBLP:journals/ai/KohaviJ97}
Kohavi~R. and John~G.~H.
\newblock Wrappers for feature subset selection.
\newblock {\em Artificial Intelligence}, 97(1-2):273--324, 1997.

\bibitem{CMB}
Kowiel~M., Brzezinski~D., Porebski~P.~J., Shabalin~I.~G., Jaskolski~M., and
  Minor~W.
\newblock Automatic recognition of ligands in electron density by machine
  learning.
\newblock {\em Bioinformatics}, page bty626, 2018.

\bibitem{FeatureSelectionACM}
Li~J., Cheng~K., Wang~S., Morstatter~F., Trevino~R.~P., Tang~J., and Liu~H.
\newblock Feature selection: A data perspective.
\newblock {\em ACM Comput. Surv.}, 50(6):94:1--94:45, Dec. 2017.

\bibitem{Li2017}
Li~Y., Li~T., and Liu~H.
\newblock Recent advances in feature selection and its applications.
\newblock {\em Knowledge and Information Systems}, pages 1--27, 2017.

\bibitem{GoldenRatio}
Livio~M.
\newblock {\em The Golden Ratio: The Story of Phi, the World's Most Astonishing
  Number}.
\newblock Broadway Books, 2002.

\bibitem{journals/ker/MariscalMF10}
Mariscal~G., Marban~O., and Fernandez~C.
\newblock A survey of data mining and knowledge discovery process models and
  methodologies.
\newblock {\em Knowledge Engineering Review}, 25(2):137--166, 2010.

\bibitem{scikit-learn}
Pedregosa~F. and et~al.
\newblock Scikit-learn: Machine learning in {P}ython.
\newblock {\em Journal of Machine Learning Research}, 12:2825--2830, 2011.

\bibitem{Perrakis1999}
Perrakis~A., Morris~R., and Lamzin~V.~S.
\newblock Automated protein model building combined with iterative structure
  refinement.
\newblock {\em Nature Structural Biology}, 6:458--463, 1999.

\bibitem{Pozharski2013}
Pozharski~E., Weichenberger~C.~X., and Rupp~B.
\newblock {Techniques, tools and best practices for ligand electron-density
  analysis and results from their application to deposited crystal structures}.
\newblock {\em Acta Crystallographica Section D}, 69(2):150--167, Feb 2013.

\bibitem{Sayes2007}
Saeys~Y., Inza~I., and Larrañaga~P.
\newblock A review of feature selection techniques in bioinformatics.
\newblock {\em Bioinformatics}, 23(19):2507--2517, 2007.

\bibitem{Sommer2007}
Sommer~I., Müller~O., Domingues~F.~S., Sander~O., Weickert~J., and Lengauer~T.
\newblock Moment invariants as shape recognition technique for comparing
  protein binding sites.
\newblock {\em Bioinformatics}, 23(23):3139--3146, 2007.

\bibitem{RESOLVE}
Terwilliger~T.~C.
\newblock Solve and resolve: Automated structure solution and density
  modification.
\newblock In {\em Macromolecular Crystallography, Part D}, volume 374 of {\em
  Methods in Enzymology}, pages 22 -- 37. Academic Press, 2003.

\end{thebibliography}
\end{document}